\documentclass[a4paper,fleqn]{cas-dc}

\usepackage{threeparttable}
\usepackage{bbding}
\usepackage{makecell}
\usepackage{subfigure}
\usepackage[authoryear]{natbib}
\usepackage {graphicx} 
\usepackage{hyperref}
\hypersetup{
	colorlinks=true,
	linkcolor=red,
	filecolor=blue,      
	urlcolor=Fuchsia,
	citecolor=Cerulean,
}
\usepackage{tikz}

\usepackage{gensymb}
\usepackage{lineno}

\usepackage{caption}

\def\tsc#1{\csdef{#1}{\textsc{\lowercase{#1}}\xspace}}
\tsc{WGM}
\tsc{QE}

\begin{document}
\let\WriteBookmarks\relax
\def\floatpagepagefraction{1}
\def\textpagefraction{.001}

\shorttitle{}    

\shortauthors{}  

\title [mode = title]{A Large-Scale Remote Sensing Dataset and VLM-based Algorithm for Fine-Grained Road Hierarchy Classification} 


\tnotetext[1]{This work was supported by the National Natural Science Foundation of China under Project 42371343, and Guangdong Basic and Applied Basic Research Foundation with Grant No.2024A1515010986.} 

\author[1]{Ting Han}
\cormark[1]
\author[1,2]{Xiangyi Xie}
\cormark[1]
\author[1]{Yiping Chen}[orcid=0000-0003-1465-6599]
\cormark[2]
\ead{chenyp79@mail.sysu.edu.cn}
\author[1]{Yumeng Du}
\author[1]{Jin Ma}
\author[3]{Aiguang Li}
\author[4]{Jiaan Liu}
\author[5]{Yin Gao}
\cormark[2]
\ead{gaoyin@ngcc.cn}

\affiliation[1]{organization={School of Geospatial Engineering and Science},
            addressline={Sun Yat-Sen University}, 
            city={Zhuhai},
            postcode={519082}, 
            country={China}}

\affiliation[2]{organization={Transport Bureau of Nanhai District}, 
	        city={Foshan},
	        postcode={528200}, 
	        country={China}}

\affiliation[3]{organization={Laboratory of Intelligent Collaborative Computing},
	        addressline={University of Electronic Science and Technology of China}, 
	        city={Chengdu},
	        postcode={611731}, 
	        country={China}}

\affiliation[4]{organization={United Nations University Institute in Macau}, 
            city={Macau S.A.R.},
            country={China}}

\affiliation[4]{organization={Moganshan Geospatial Information Laboratory}, 
            city={Deqing},
            \postcode={313200},
            country={China}}

\cortext[1]{The authors contribute equally.}
\cortext[2]{Corresponding author}

\begin{abstract}
Accurate and up-to-date road hierarchy information is fundamental to transport infrastructure mapping, road asset inventory, maintenance planning, and digital urban management. Although high-resolution remote sensing imagery enables large-area road mapping, existing datasets and methods are still poorly aligned with practical infrastructure needs, as most benchmarks focus on binary road extraction and provide limited support for hierarchy-aware mapping. This gap restricts the development of automated workflows that jointly support road surface delineation, road network reconstruction, and fine-grained road hierarchy classification. In this work, we present \textbf{SYSU-HiRoads}, a large-scale hierarchical road dataset, and \textbf{RoadReasoner}, a vision-language-geometry framework for automatic multi-grade road mapping from remote sensing imagery. SYSU-HiRoads is built from GF-2 imagery covering 3,631 km$^{2}$ in Henan Province, China, and contains 1,079 image tiles at 0.8 m spatial resolution. Each tile is annotated with dense road masks, vectorized centerlines, and three-level hierarchy labels, enabling the joint training and evaluation of segmentation, topology reconstruction, and hierarchy classification. Building on this dataset, RoadReasoner is designed to generate robust road surface masks, topology-preserving road networks, and semantically coherent hierarchy assignments. We strengthen road feature representation and network connectivity by explicitly enhancing frequency-sensitive cues and multi-scale context. Moreover, we perform hierarchy inference at the skeleton-segment level with geometric descriptors and geometry-aware textual prompts, queried by vision–language models to obtain linguistically interpretable grade decisions. Experiments on SYSU-HiRoads and the CHN6-CUG dataset show that RoadReasoner surpasses state-of-the-art road extraction baselines and produces accurate and semantically consistent road hierarchy maps with 72.6\% OA, 64.2\% F1 score, and 60.6\% SegAcc. The dataset and code will be publicly released to support automated transport infrastructure mapping, road inventory updating, and broader infrastructure management applications.
\end{abstract}


\maketitle


\section{Introduction}

Roads are fundamental components of transport infrastructure and play a central role in urban mobility, logistics, infrastructure maintenance, and regional development \citep{chen2022road,zhang2026multimodal,pena2026bim}. Accurate and up-to-date road information is essential for infrastructure inventory, transport planning, traffic management, and digital urban governance. With the rapid development of high-resolution Earth observation platforms and sensing technologies, remote sensing imagery has become an efficient data source for large-scale and automated road mapping. As a result, road extraction from high-resolution remote sensing images has emerged as a key task in remote sensing and geospatial analysis, with broad application potential in infrastructure monitoring, urban planning, and intelligent transportation systems \citep{Zhang2018RoadEB, Chen2023SemiRoadExNet, han2024epurate, han2025cityinsight,gao2025identifying}. 

With the rapid advancement of high-resolution Earth observation systems, including the Gaofen series, WorldView satellites, and commercial aerial imaging platforms, detailed and large-scale observations of road networks have become increasingly accessible. Consequently, the key challenge is no longer the availability of imagery, but how to automatically and reliably extract, reconstruct, and interpret complex road networks from massive image data for infrastructure mapping and management \citep{sun2026text}.

Early studies mainly relied on handcrafted features and classical image processing or probabilistic models to delineate roads from remote sensing imagery. Morphological operators, template matching, active contours, and Markov random fields were designed to exploit the elongated, relatively homogeneous, and topologically connected nature of roads \citep{li2017recognition,Movaghati2010Road,LI2016Region,Chaudhuri2012Semi,Grinias2016MRFbasedSA,ALSHEHHI2017Hierarchical}. These approaches can be effective in constrained scenarios, but their performance is highly sensitive to parameter settings and background clutter, and their ability to generalize to diverse environments (e.g., dense urban cores, industrial areas, mountainous regions) is limited. In particular, occlusions by trees and buildings, shadows, heterogeneous pavements, and complex junctions often lead to fragmented or missing roads in the extracted maps.

The advent of deep convolutional neural networks (CNNs) and more recent Transformer-based architectures \citep{kyem2026self,guo2023pavement,zhang2023d} has led to substantial progress in road extraction. These methods have shown a strong capability in learning discriminative features directly from data and in improving the completeness of road masks \citep{Zhang2018Road,Ronneberger2015U-Net,Zhou2018DLinkNetLW,Campos2020Deep,Yang2023GeoscienceR,Zhang2022DCS-TransUperNet}. Subsequent work further incorporated dilated convolutions, attention mechanisms, and auxiliary losses on boundaries or centerlines to enhance the continuity and topology of road predictions.

Despite these advances, three important limitations remain: (1) Pixel-wise segmentation methods often produce geometrically noisy or locally inconsistent masks, making it difficult to derive high-quality vector road networks without additional post-processing. (2) Existing models rarely account for the semantic relationship between road geometry (e.g., width, length, and curvature) and its hierarchical role in the transport infrastructure system. (3) Most methods focus on binary road extraction and treat all roads as a single semantic class, ignoring their hierarchical attributes such as functional class, capacity, and administrative level, which are crucial for many downstream tasks (e.g., speed estimation, accessibility analysis, infrastructure assessment, and route planning).

The lack of suitable benchmark datasets is a major bottleneck behind these limitations. Most publicly available road datasets derived from remote sensing imagery, such as Massachusetts, DeepGlobe, and CHN6-CUG \citep{MnihThesis,demir2018deepglobe,zhu2021global}, provide only pixel-level binary masks without any information on road hierarchy or matched vector-level networks \citep{abdollahi2021roadvecnet}. In contrast, road-network-oriented datasets, such as SpaceNet Roads, RoadTracer, and City-scale \citep{van2018spacenet,Bastani2018RoadTracerAE,He2020Sat2GraphRG}, mainly offer vectorized centerlines that are rasterized with fixed widths; they lack true pixel-wise masks and do not encode multi-grade attributes of road infrastructures. This separation between segmentation-oriented and network-oriented datasets makes it difficult to jointly study road surface extraction and network reconstruction within a unified framework. 

Moreover, accessible, high-quality road data in China is particularly scarce. OpenStreetMap (OSM) data, which are widely used in many existing benchmarks, exhibit non-negligible offsets in China due to coordinate restrictions and thus cannot be directly used as precise labels \citep{ZHU2021353}. To the best of our knowledge, there is still no large-scale, high-resolution dataset that simultaneously provides aligned pixel-level masks, vector-level centerlines, and fine-grained hierarchical road labels for Chinese cities.

At the same time, the rapid development of vision language models (VLMs) \citep{radford2021learning} and multimodal generative models \citep{achiam2023gpt,mao2024gpteval}, such as GPT-v, has opened up new opportunities for infrastructure analysis \citep{xiao2025llm,tao2025advancements, silva2024multilingual,zavras2025mind,han2025cityinsight}. These foundation models are pre-trained on massive image–text pairs and encode rich semantic knowledge about objects, scenes, and their functional roles \citep{li2023rs}. Recent works have begun to explore VLMs for land-cover classification \citep{weng2025vision,wu2025farmseg_vlm}, change detection \citep{li2024semicd}, and cross-modal retrieval \citep{danish2025geobench}, but their potential for modeling structured geographic entities, such as road networks, particularly for inferring road hierarchy from geometric and contextual cues, remains largely unexplored. A key observation of our work is that translating quantitative geometric attributes (e.g., width, length, and straightness) and engineering expertise \citep{jiang2022building} into language-compatible descriptions with visual features is effective for producing reliable, interpretable road-grade predictions.

To address these gaps, we present \textbf{SYSU-HiRoads}, a new large-scale hierarchical road dataset, together with \textbf{RoadReasoner}, a language–vision–geometry framework for automatic road extraction and fine-grained hierarchy classification from high-resolution remote sensing imagery. SYSU-HiRoads provides (1) pixel-level road masks, (2) vector-level road centerlines, and (3) three-level hierarchical labels (high, medium, and low grades). This design allows researchers to study road surface segmentation and network reconstruction within a single unified benchmark and to explicitly link geometric properties with hierarchical attributes. Building on SYSU-HiRoads, we further propose RoadReasoner, which couples a carefully designed road extractor (FORCE-Net) with a text-guided hierarchical road network grader (T-HRN) for road segmentation, network reconstruction, and road hierarchy grade assignments. The main contributions are as follows:
\begin{itemize}
    \item We construct \textbf{SYSU-HiRoads}, the first large-scale hierarchical road benchmark over Chinese cities, to the best of our knowledge, that jointly provides aligned pixel-level masks, vector-level centerlines, and three-levels of road grades.
    \item We propose RoadReasoner, which consists of FORCE-Net and T-HRN, using VLM and human-interpretable prompts for automatic road extraction and road hierarchy classification. 
    \item We conduct extensive experiments on SYSU-HiRoads and additional benchmarks, demonstrating that our framework not only improves road segmentation accuracy and connectivity but also delivers reliable fine-grained hierarchy classification in transport infrastructure mapping.
\end{itemize}

\section{Previous Works}

\begin{table*}[]
     \centering
     \caption{Summary of existing datasets for road extraction and road network extraction from remote sensing images, compared to our proposed SYSU-HiRoads dataset. SYSU-HiRoads provides aligned pixel-level and vector-level annotations, along with multi-grade hierarchical road attributes.}
     \resizebox{\textwidth}{!}{
     \begin{tabular}{c|c|c|c|c|c|c|c|c}
         \toprule
         Dataset & Year & Source & Resolution & Image Size & Data Volume & Task & Multi-Grade & Access. \\
         \midrule
         Massachusetts \citep{MnihThesis}& 2013 & Aerial Image & 1 m & $ 1,500 \times 1,500 $ & 1,771 & Road Extraction & \XSolidBrush & \Checkmark \\
         CasNet \citep{cheng2017automatic} & 2017 & Google Earth & 1.2 m &$ 600 \times 600 $& 224 & Road Extraction & \XSolidBrush & \XSolidBrush \\
         DeepGlobe \citep{demir2018deepglobe} & 2018 & DigitalGlobe +Vivid & 0.5 m & $ 1,024 \times 1,024 $ & 6,226 & Road Extraction & \XSolidBrush & \XSolidBrush \\
         CHN6-CUG \citep{zhu2021global} & 2021 & Google Earth & 0.5 m &$ 512 \times 512 $ & 4,511 &Road Extraction & \XSolidBrush & \Checkmark \\
         LRSNY \citep{chen2021reconstruction} & 2021 & - & 0.5 m &$ 1,000 \times 1,000 $& 1,368 & Road Extraction & \XSolidBrush & \Checkmark \\
         Gansu Mountain \citep{zhou2021split} & 2021 & ZY-3 & 2 m & $ 256 \times 256 $ & 255 & Road Extraction & \XSolidBrush & \Checkmark \\
         GlobalRoadSet \citep{Lu2024GlobalRE}& 2024 & Google Earth & 1 m &$ 1,024 \times 1,024 $& 47,210 & Road Extraction & \XSolidBrush & \Checkmark \\
         \midrule
         SpaceNet \citep{van2018spacenet} & 2018 & WorldView-3 & 1 m& $ 400 \times 400 $& 2,551 & Road Network & \XSolidBrush & \Checkmark \\
         RoadTracer \citep{Bastani2018RoadTracerAE}& 2018 & Google Earth & 0.6 m & $ 4,096 \times 4,096 $& 300 & Road Network & \XSolidBrush & \Checkmark \\
         City-scale \citep{He2020Sat2GraphRG}& 2020 &Google Map & 1 m &$ 2,048 \times 2,048 $ & 180 & Road Network & \XSolidBrush & \Checkmark\\%
         \midrule
         \rowcolor{gray!20}
         \textbf{SYSU-HiRoads (Ours)} & 2025 & GF-2 & 0.8 m & $ 1,024 \times 1,024 $ & 1,079 & \makecell{Road Extraction \\ Road Network} & \Checkmark & \Checkmark \\
         \bottomrule
     \end{tabular}
     }
     \label{existing road dataset}
 \end{table*}

\subsection{Road Extraction Methods}

Roads generally exhibit a relatively uniform width with minimal variation \citep{ALSHEHHI2017Hierarchical}, maintaining a slender and uniform structure. Roads exhibit stable shape and appearance characteristics. Previous studies have demonstrated that the design, adjustment, and optimization of handcrafted features based on the morphological properties of the road can effectively facilitate road extraction tasks in various scenarios \citep{Movaghati2010Road,LI2016Region,COURTRAI2016195}.

Chaudhuri et al. \citep{Chaudhuri2012Semi} applied a semi-automatic morphological method to extract road centerlines. However, this method only considers the morphological characteristics of the roads and neglects their topological structure. Bae et al. \citep{Bae2015Automatic} proposed an algorithm based on features such as road width, length, contrast, and orientation, which simultaneously extracts both road centerlines and road surface masks. Grinias et al. \citep{Grinias2016MRFbasedSA} utilized morphological features along with a Markov random field model to achieve unsupervised road segmentation. Building on this, Alshehhi et al. \citep{ALSHEHHI2017Hierarchical} introduced a directional morphological operator by incorporating path opening and path closing operators into road extraction from high-resolution imagery. Although this method improves the connectivity of the results, it is susceptible to cumulative errors and produces un-smooth segmentation results. Wegner et al. \citep{WEGNER2015Road} employed a probabilistic representation of road network pixels to classify road categories and then combined color and texture features to perform a least-cost path search within the network. However, this approach is sensitive to threshold parameters and is not suitable for complex variable scenarios.

Traditional methods can effectively extract features such as road shape, width, and orientation, but this advantage is based on manually setting parameters. Recently, deep learning-based methods have demonstrated significant success in road extraction tasks due to their ability to automatically learn features and their robust learning capabilities. \cite{Zhang2018Road} were the first to apply U-Net \citep{Ronneberger2015U-Net} to aerial image road extraction. To address connectivity issues and enhance topological relationships, \cite{Zhou2018DLinkNetLW} incorporated parallel atrous convolutions into LinkNet to expand the receptive field, with \cite{Campos2020Deep} further refining this approach. \cite{Woo2018LinkNetRE} adopted a strategy of sharing module information during encoder design, while \cite{Marmanis2018Classification} improved a fully convolutional neural network by integrating class boundary information during training. Although these strategies significantly enhanced road continuity, they still faced challenges in accurately segmenting small-scale structures. 

To fully exploit fine-grained textures and deep semantic features, \cite{Yang2023GeoscienceR} proposed a multi-scale road context extraction module that effectively captures long-range dependencies. \cite{Zhang2022DCS-TransUperNet} utilized a dual-resolution Transformer module along with feature fusion to acquire global contextual information. However, due to the inherent limitations of convolutional operators, these methods still cannot effectively capture the global features of roads. \cite{Sun2019Leveraging} designed a 1D transposed convolution that integrates crowdsourced GPS data and aerial imagery for road extraction. \cite{Mei2021CoANet} designed a stripe-shaped convolution to capture remote contextual relationships.  

\subsection{Road Datasets}

Publicly available road datasets are shown in Tab.~\ref{existing road dataset}.

The Massachusetts Road dataset \citep{MnihThesis} contains 1,171 optical remote sensing images, each with dimensions of $ 1,500 \times 1,500 $ pixels and a spatial resolution of approximately 1 meter per pixel. This dataset covers urban, suburban, and rural environments with binary mask annotations. The DeepGlobe Road dataset \citep{demir2018deepglobe} comprises 6,226 images for training, 1,243 images for validation, and 1,101 images for testing. Each image has a size of $ 1,024 \times 1,024 $ and a resolution of 0.5 m. Pixel-level road annotations are provided for each image.

\begin{figure*}[!t]
    \centering
    \includegraphics[width=0.85\linewidth]{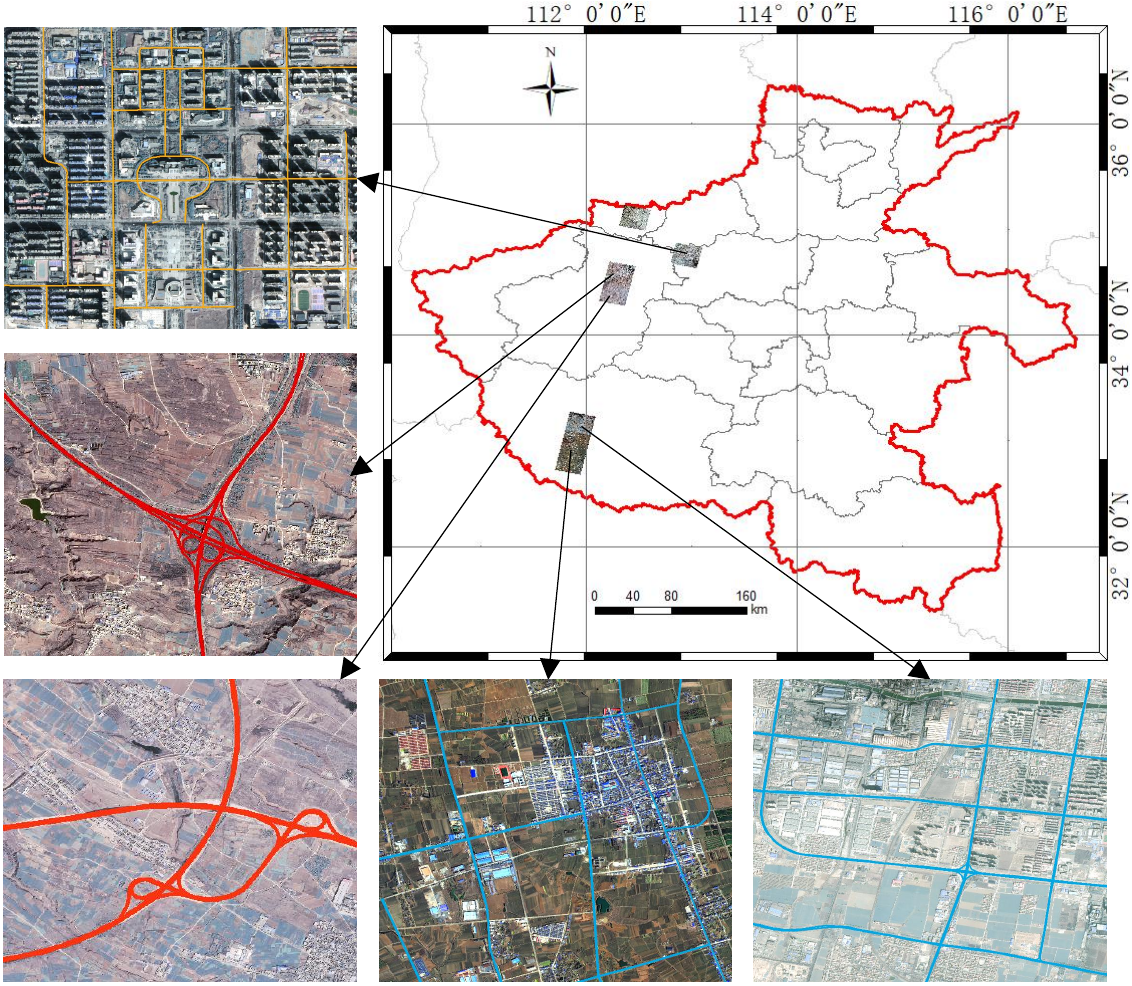}
    \caption{Coverage Schematic of SYSU-HiRoads Dataset, which covers 3,631 km$^{2}$ of land in Henan, China. We use different colors to mark roads of various levels, with red, blue, and yellow representing high, medium, and low levels, respectively.}%
    \label{fig: Coverage Schematic of Multi-Grade Road Dataset}
\end{figure*}

The RoadTracer dataset \citep{Bastani2018RoadTracerAE} provides road centerline annotations without pixel-level segmentation labels, which limits its suitability for pixel-level road segmentation tasks. This constraint significantly restricts the dataset's applicability to segmentation tasks. The City-scale dataset \citep{He2020Sat2GraphRG} comprises 180 satellite images sourced from Google Maps, each having a size of $2,048 \times 2,048$ pixels and a spatial resolution of 1 meter per pixel. The dataset provides both vector-level road network annotations and corresponding segmentation maps. However, the provided segmentation maps are not true pixel-level annotations. Instead, they are generated directly by rasterizing the road centerlines, with all roads uniformly represented by a fixed width of three pixels.

The CasNet dataset \citep{cheng2017automatic} consists of images with a spatial resolution of 1.2 meters per pixel. Each image in the dataset has dimensions of $600 \times 600$ pixels, and the annotations include both the road regions and the corresponding road centerlines. The CHN6-CUG Road dataset \citep{zhu2021global} is a dataset consisting of road images with pixel-level annotations and a spatial resolution of 0.6 meters. The dataset covers diverse urban areas in multiple major Chinese cities, including Beijing, Shanghai, and Wuhan, providing comprehensive geographic coverage. Specifically, it offers detailed annotations of road networks, making it suitable for evaluating pixel-level road extraction methods in various urban environments.

The SpaceNet Roads dataset \citep{van2018spacenet} provides road centerline annotations collected from four cities: Las Vegas, Paris, Shanghai, and Khartoum. The dataset comprises 2,780 images paired with the corresponding road centerline maps. Each image has a spatial resolution of 0.3 m and an image size of $1,300 \times 1,300$ pixels. Although the dataset categorizes roads into five different types, it does not include detailed annotations distinguishing these categories.

\section{SYSU-HiRoads}

\subsection{Challenges of Existing Datasets}

\textbf{Existing road datasets cannot simultaneously meet the requirements of multiple tasks.} Currently, datasets like the DeepGlobe and Massachusetts Road Dataset lack road vectors and can only perform semantic segmentation by classifying pixels as roads or background (non-road). On the other hand, road network datasets such as RoadTracer and the City-scale dataset lack pixel-level annotations. These datasets are insufficient for road segmentation tasks. This limitation makes it challenging to perform both road segmentation and road network extraction within a single framework.

\textbf{Accessible road data in China is scarce.} The publicly available OpenStreetMap exhibits a significant offset in the Chinese region, making it unsuitable for direct use in generating road datasets. In addition, given the scarcity of road data and the lack of geographic information in China, it poses a significant challenge for the extraction and analysis of road networks in the region \citep{ZHU2021353}.

\textbf{Previous road datasets lack hierarchical road labeling.} Current road datasets focus primarily on road semantics without providing detailed information on road hierarchical classification. However, in reality, roads have various attributes related to their role, nature, and importance. According to different standards and regulations, roads can be classified into different grades. A small number of categories is insufficient to represent the full range of road attributes, while an excessive number of categories can complicate data processing and applications. Determining the appropriate classification and organization of roads is also a significant challenge.

\subsection{Description of SYSU-HiRoads}

\begin{figure*}[!t]
     \centering
     \includegraphics[width=\linewidth]{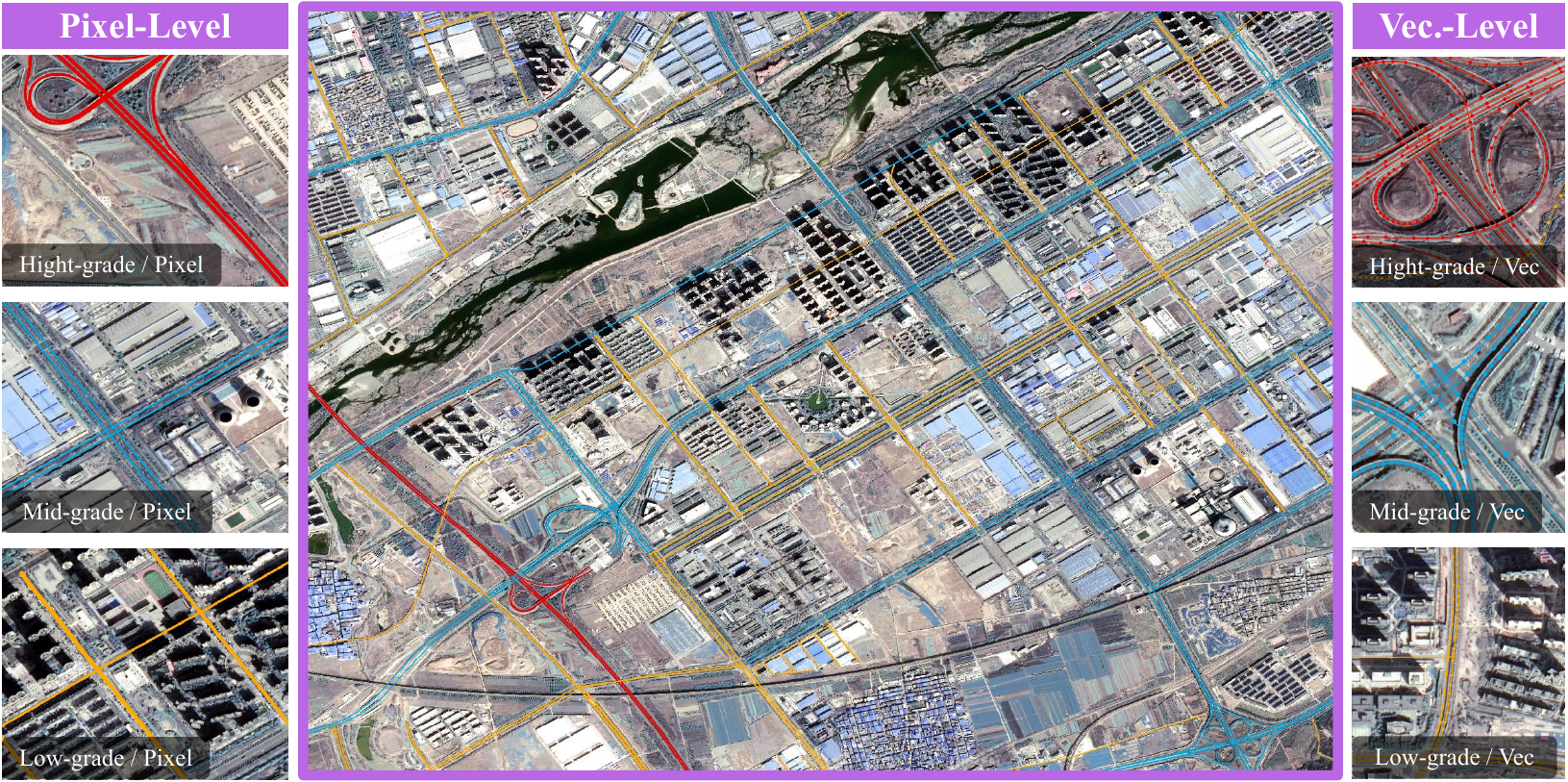}
     \caption{Representative examples of the SYSU-HiRoads Dataset, which contains road pixel-level and vector-level annotations. We use different colors to mark roads of various levels, with red, blue, and yellow representing high, medium, and low levels, respectively.}
     \label{fig: Dataset Diagram}
\end{figure*}

SYSU-HiRoads includes various cities in Henan, China, as shown in Fig.~\ref{fig: Coverage Schematic of Multi-Grade Road Dataset}, such as Luoyang, Jiyuan, and Zhengzhou, representing various levels of development. The dataset contains various types of roads, including highways, national roads, provincial roads, county roads, and more. It covers a wide range of scenes, including urban roads, factory roads, overpasses, interchanges, and roundabouts.

The original image is a 0.8 m high-resolution remote sensing image from the Gaofen-2 Satellite (GF-2), covering 3,631 km$^{2}$ and has a resolution of 0.8 meters. We cut the original large image into pieces with a size of $ 1,024 \times 1,024 $, generating 1,079 images. These images are further divided into training, validation, and test sets, containing 871, 108, and 100 images, respectively.

\subsection{Annotation \& Marking Guide}

We manually label the ground truth of the road areas in the original large image and implement the division in the same manner as in the original image. Thus, there is a one-to-one correspondence between the original image pieces and the ground truth label pieces.

\textbf{Road Selection.} We take into account factors such as the clarity of roads in satellite imagery and actual traffic flow, and we choose to retain various major road types (national highways, provincial highways, and county roads) while reducing the selection of other narrow or less significant road types (extremely narrow paths reserved for pedestrians or non-motorized vehicles, as well as temporary construction access roads, etc.). This approach minimizes the negative impact of uncertainty, ensuring that the selected road types are more representative and valuable for research.

\textbf{Road Hierarchical Classification.} We determine the level and type of the road based on the administrative hierarchy, usage function, and traffic flow factors\footnote{https://www.mohurd.gov.cn/}. This process primarily combines OSM data with the Chinese Road Administrative Classification System to identify whether a road belongs to a high-level road (including highways and expressways), a medium-level road (mainly national highways, provincial highways, and some first-class and second-class roads), or a low-level road (mainly county roads, township roads, and third-class roads), as shown in Fig.~\ref{fig: Dataset Diagram}.

\textbf{Road Centerline Drawing.} When drawing the road centerline, if the road has clear bidirectional lanes with a distinct median, we draw the centerlines for each direction separately. Otherwise, we draw a single line along the center of the road. We then assign the appropriate level labels to each marked road area based on the actual conditions of the road.

\textbf{Road polygon generation.} Road widths have specific ranges and standards. We obtain data indicating that (1) the typical width of a single lane on a highway is 3.75 meters, with a total of 4–8 lanes; (2) urban arterials generally have a single lane width of 3.5 to 3.75 meters, with 4-8 lanes; (3) secondary roads have a lane width of approximately 3.25 to 3.5 meters, with 2 or 4 lanes. Based on the road centerlines, we generate road buffers. Specifically, from the extracted or drawn road centerlines, we create multiple buffers of different widths according to the common width values of the roads. Then, based on the actual visual width of the road in the imagery, the relationship to the surrounding environment, and the characteristics of the road classification, we analyze each road and select the buffer that best fits the actual width of the road. Finally, the chosen buffer is converted into a polygon to delineate the road area.

\textbf{Data organization and storage.} We store the annotated road centerlines following a specific directory structure and save them in JSON format. Finally, the road surface polygons are converted into raster data in TIFF format for convenient data management and usage. Two professional researchers perform the annotations, followed by a secondary review to ensure the accuracy of the annotations.

\subsection{Advantages}

\begin{itemize}
    \item \textbf{Possessing more detailed annotation information:} The SYSU-HiRoads Dataset classifies roads into three levels based on administrative levels, traffic flow, etc. It can assist in accurately predicting vehicle speeds and route planning.
    
    \item \textbf{Capable of meeting various task requirements:} Our dataset combines pixel-level annotation with road network information. Existing datasets either have only pixel-level road surface labels or only vectorized road centerline information. This dataset can meet the requirements for both road segmentation and road network extraction method research simultaneously.
    
    \item \textbf{Rich in image detail information:} This dataset has a high resolution. The image data are sourced from the GF-2 satellite data, with an image resolution of 0.8 m per pixel and an image size of $ 1,024 \times 1,024 $. It can present more road details and meet the needs of road research.
\end{itemize}

\section{RoadReasoner}

\begin{figure*}[!t]
    \centering
    \includegraphics[width=\linewidth]{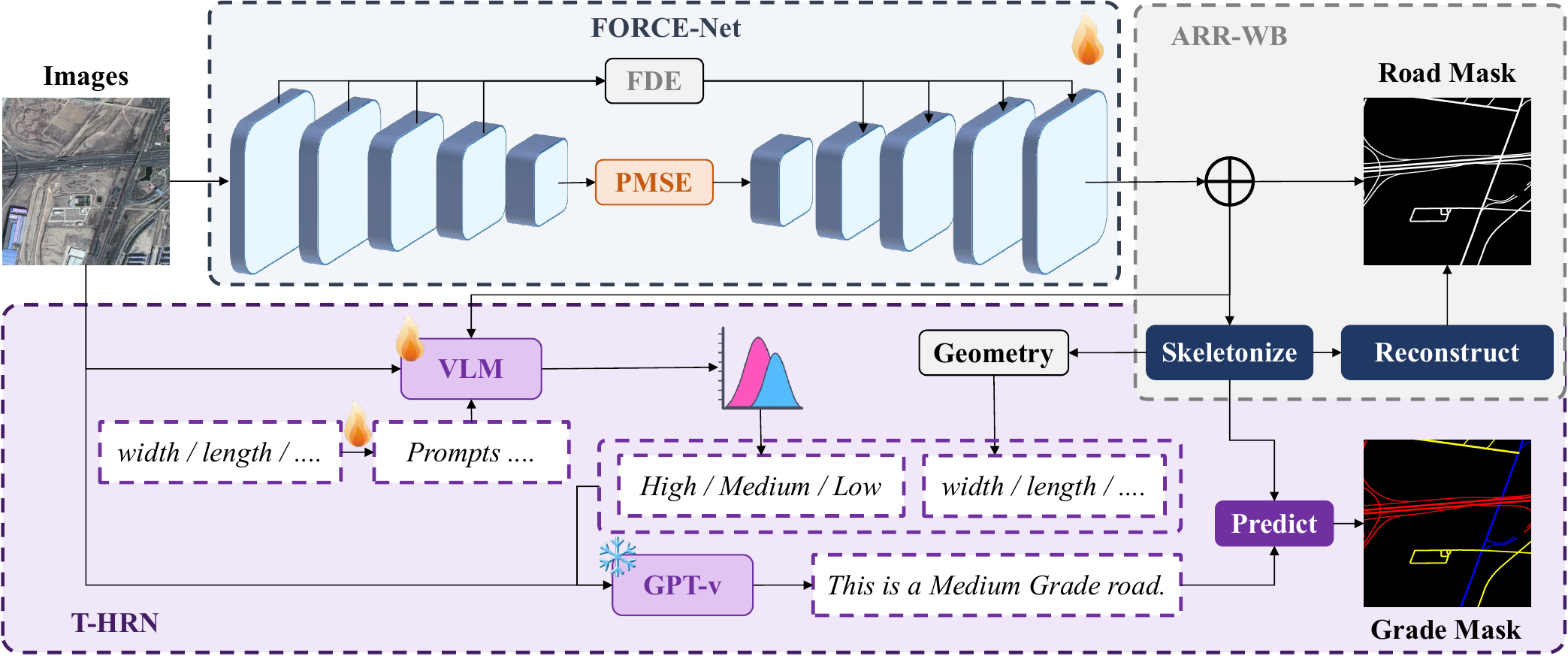}
    \caption{Overview of the proposed \textbf{RoadReasoner} framework. Given an input remote sensing image, the upper branch adopts FORCE-Net with the FDE and PMSE modules to extract a complete road mask within the ARR-WB scheme. The lower T-HRN branch receives the extracted road skeleton together with geometric attributes and converts them into textual prompts for a vision–language model (VLM), which infer a linguistic description of the road grade. The predicted grades are then fused with the reconstructed road mask to generate the final grade mask, where roads are color-coded according to their hierarchical classes.}
    \label{fig: Overall network structure}
\end{figure*}

An overview of the proposed RoadReasoner framework is shown in Fig.~\ref{fig: Overall network structure}. RoadReasoner consists of two main parts: (1) a road extraction network, \textbf{FORCE-Net}, which employs the Frequency Domain Enhancement (FDE) block and the Parallel Multi-Scale Feature Extraction (PMSE) block to generate a complete road mask and integrates the ARR-WB module for multi-weighted backtracking–based reconstruction of broken road centerlines; (2) a text-guided hierarchical road network grader, \textbf{T-HRN}, which takes the reconstructed road skeleton and its geometric attributes, converts them into language prompts for a VLM and GPT-v, and predicts a grade mask where roads are assigned to different hierarchical classes. We detail the designs of these two components in the following subsections.

\subsection{FORCE-Net}

We use a reduced version of our proposed RSDMNet \citep{xie} as the backbone to construct an end-to-end feature extractor. The encoder consists of an initial layer and four downsampling layers to extract hierarchical features, represented by $x_{1} - x_{5}$ with dimensions of $[64, 64, 128, 256, 512]$. In the decoder, bilinear interpolation is applied to upsample the features, obtaining $x_{1}' - x_{4}'$ corresponding to the features of the encoder. We use PMSE as a bridge between the encoder and the decoder to obtain $ x_{5}'$, facilitating the extraction of sufficient global contextual information. Moreover, to capture more effective global road information, we design Fourier-based FDE blocks. The hierarchical features of the encoder are fused and integrated with the decoder features after applying Fourier attention. Finally, the output from the decoder passes through an activation function to generate the final road prediction, which is supervised using binary cross-entropy loss.

\begin{figure}[]
    \centering
    \includegraphics[width=0.9\linewidth]{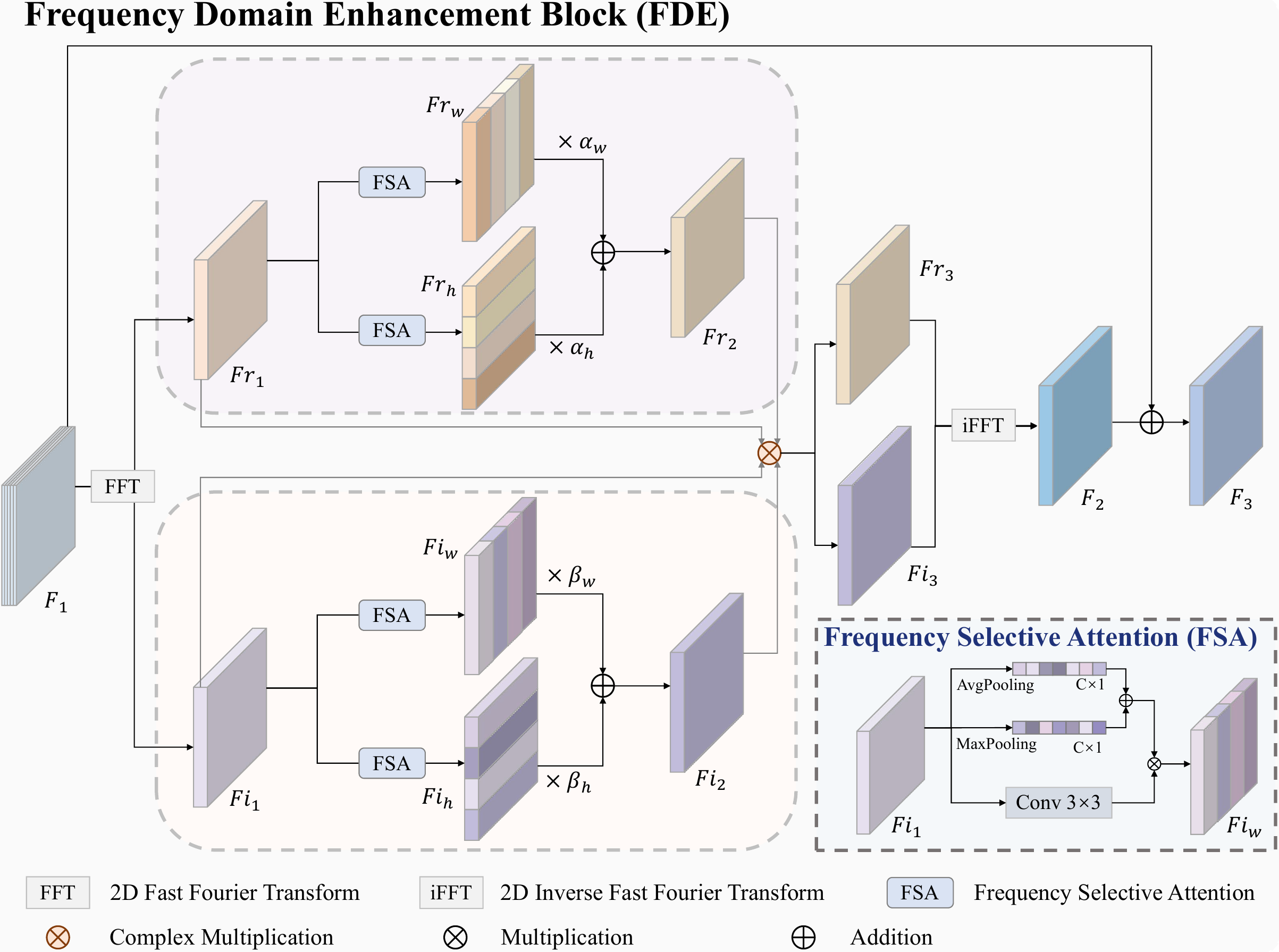}
    \caption{Illustration of Frequency Domain Enhancement Block.}
    \label{fig: Frequency Domain Enhancement Block}
\end{figure}

\subsubsection{Frequency Domain Enhancement}

The Fourier transform is an operator that directly acts on the spatial dimensions of images, analyzing and enhancing them from the perspective of frequency to capture global features. We observed that the low-frequency components after the Fourier transformation represent global structural information, such as road surfaces. In addition, the high-frequency information in the image is mainly used to represent details of edges and textures, such as the edges of roads. Therefore, combining these components can help improve the representation of both global information and spatial features of the image. 

Inspired by Fourier neural operators \citep{Li2020FourierNO}, we design a Fourier-based feature enhancement module, as shown in Fig.~\ref{fig: Frequency Domain Enhancement Block}. Firstly, we fuse the hierarchical features as $F_{1}$. After being transformed by the Fast Fourier Transform (FFT), $F_{1}$ is decomposed into low-frequency features $F_{r1}$ and high-frequency features $F_{i1}$, which are then fed into parallel dual-branch modules. These modules share the same structure: we use Frequency Selective Attention (FSA) to extract global information in the horizontal and vertical directions and then multiply it with learnable parameters ($\alpha_{w}, \alpha_{h}, \beta_{w}, \beta_{h}$) to obtain the enhanced features $F_{r2}$ and $F_{i2}$, where $\alpha$ and $\beta$ act on $F_{r1}$ and $F_{i1}$, and $w$ and $h$ denote the vertical and horizontal directions, respectively.

Subsequently, the enhanced features are reshaped into new Fourier frequency-domain features and multiplied with the original Fourier features in a complex manner. We consider that information transmission in the frequency domain enables the capture of richer features while retaining the original characteristics. Finally, we obtain the enhanced feature $F_{3}$ by adding the residual of the inverse Fast Fourier Transform feature $F_{3}$ and the input feature $F_{1}$.

\subsubsection{Parallel Multi-Scale Feature Extraction}

\begin{figure}[]
    \centering
    \includegraphics[width=0.9\linewidth]{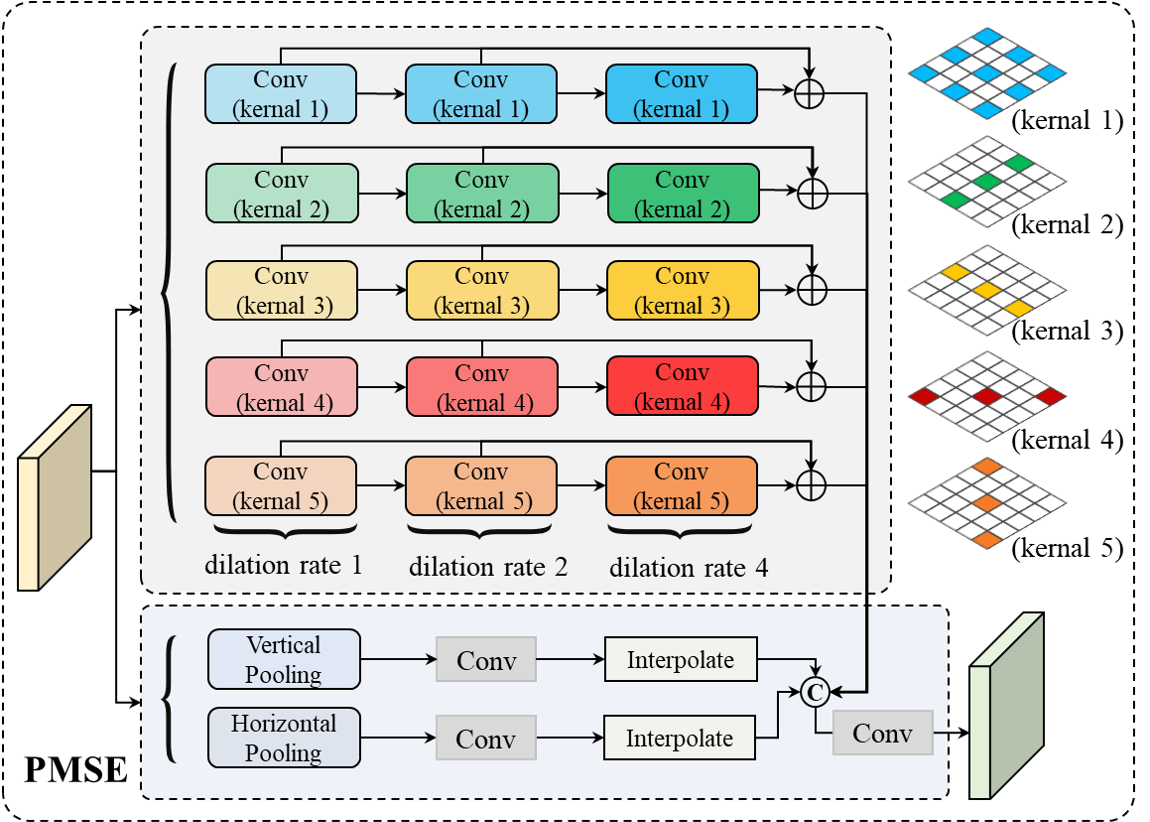}
    \caption{Illustration of Parallel Multi-Scale Feature Extraction Block.}
    \label{fig: Parallel Multi-Scale Feature Extraction Block}
\end{figure}

We introduce the PMSE Block to enhance edge information and road connectivity, as illustrated in Fig.~\ref{fig: Parallel Multi-Scale Feature Extraction Block}. The PMSE utilizes five parallel dilated convolutions to capture multi-scale contextual features. These dilated convolution kernels operate in four directions with dilation rates of ${1, 2, 4}$, which enables them to better adapt to the shape of roads. Simultaneously, we employ $3 \times 3$ convolutions, horizontal pooling, and vertical pooling to extract features at different scales and orientations while preserving the global road information. The features of the parallel frameworks are then fused and combined with the original input features.

\subsubsection{Automatic Road Reconstruction}

To obtain a more complete road mask and road centerlines, we have developed an automatic reconstruction method to complete the road centerline. Firstly, we detect the road endpoints, and then a multiple backtracking weighted average method is used to determine the orientation angles of the endpoints. Next, we use the orientation angles of the endpoints, the orientation angles of the lines connecting the endpoints, and the distances between the endpoints as constraints to calculate the scores of matching between the endpoints in order to select the best connection point \citep{wang2015auto}. Finally, the best matching points are connected.

\begin{figure}[]
    \centering
    \includegraphics[width=0.9\linewidth]{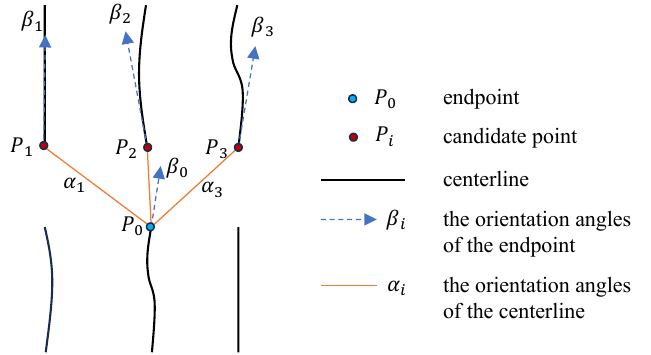}
    \caption{Illustration of multiple backtracking strategy.}
    \label{fig: multiback}
\end{figure}

We use a multiple backtracking weighted average algorithm to calculate the orientation angle. Specifically, starting from the endpoint $P_{0}$, we backtrack $l_{1}$, $l_{2}$, and $l_{3}$ pixels to obtain the backtracking points $P_{1}$, $P_{2}$, and $P_{3}$, respectively. Then, we compute the orientation angles of the lines connecting each backtracking point with the endpoints, as well as the weighted average orientation angle $\overline{\beta}$:
\begin{equation}
w_i=\frac{l_i}{\sum_{i=1}^nl_i}~~,
\end{equation}
\begin{equation}
\overline{\beta} = \frac{\sum_{i=1}^n \arctan(\frac{y_i-y_0}{x_i-x_0})\times w_i}{\sum_{i=1}^n w_i}~~,
\end{equation}
where $(x_{0}, y_{0})$ is the coordinate of $P_{0}$. We set $l_{1}$, $l_{2}$, and $l_{3}$ to 10, 15, and 20, respectively.

We then calculate the maximum orientation angle constraint based on the weighted average orientation angle. The weighted average orientation angle of endpoint $P_{0}$ reflects the general direction of the road at that endpoint. Since roads typically have small curvature, the orientation angle of the optimal connection endpoint $P'_{0}$ does not deviate significantly from the offset angle $\Delta \sigma$ of $P_{0}$. However, there may be multiple parallel roads, so we also use the orientation angle of the line connecting the endpoint and the candidate point as a constraint to avoid cross connections. As shown in Fig.~\ref{fig: multiback}, the candidate point $P_{3}$ has a smaller offset $|\beta_{3} - \beta_{0}|$ relative to $P_{0}$. The deviation between the orientation angle of the line connecting the endpoints and the weighted average orientation angle $\overline{\beta}$ of $P_{0}$, i.e., $|\alpha_{3} - \beta_{0}|$, is very large. Therefore, $P_{3}$ is not suitable as the connection point. Ultimately, we choose a maximum deviation of no more than $30^{\circ} $:
\begin{equation}
    \left\{
    \begin{aligned}
        |\beta_i-\beta_0|<30^\circ\\
        |\alpha_i-\beta_0|<30^\circ
    \end{aligned}
    \right.~~.
\end{equation}

Then, under the maximum deviation angle constraint, we use the length of the line connecting the endpoints and the directional angle deviation between the endpoints as weights to calculate the endpoint matching degree. Since the length and the directional angle deviation are measured in different units and cannot be directly weighted, we convert the directional angle deviation into an offset arc, with the endpoint as the center and the length of the connecting line as the radius, and use the arc length instead of the angle deviation for the calculation. The endpoint matching degree $D_{i}$ between endpoint $P_{0}$ and candidate point $P_{i}$ is defined as:
\begin{equation}
D_i=|\beta_i-\beta_0 |\times d(P_i,P_0)\times w_1+d(P_i,P_0)\times w_2~,
\end{equation}
where $d(\cdot)$ denotes the Euclidean distance, and $w_{1}$ and $w_{2}$ are set to 0.2 and 0.8, respectively. Finally, we select the candidate point with the best matching degree $min(D_{i})$ and connect it to the endpoint.

\subsection{T-HRN}

T-HRN is responsible for assigning hierarchical grades to the extracted road network. As shown in Fig.~\ref{fig: Overall network structure}, T-HRN operates on the reconstructed road mask produced by FORCE-Net and consists of two main parts: (1) geometry-aware prompt construction and language prior inference; and (2) road-grade prediction on the road skeleton and generation of a pixel-wise grade mask.

\subsubsection{Geometry-aware Prompt and Language Prior}

Given the reconstructed road mask, we first skeletonize the road regions and group the skeletons into individual road segments. For each segment, we compute a set of geometric attributes, including width, length, and other optional descriptors (e.g., straightness and curvature). These continuous measurements are discretized into human-interpretable categories such as “short/medium/long” and “narrow/medium/wide”. We then instantiate a collection of textual templates, where the geometric descriptors are inserted into structured prompts, e.g., “a long and wide road segment in an urban area”. Together with the corresponding image patches cropped around each road segment, the prompts are fed into a vision–language model (VLM) to obtain a geometry-aware language prior over road grades (e.g., high, medium, and low grade roads). 

In parallel, we concatenate the geometric descriptors and the VLM-derived priors and provide them, along with the image context, to GPT-v. GPT-v generates a natural-language description of the road, such as “This is a Medium Grade road.”, which serves both as an interpretable explanation and as an additional semantic constraint on the grade prediction.

\subsubsection{Road-grade Prediction}

In the second part, T-HRN aggregates the multi-source cues from geometry, VLM, and GPT-v to predict a discrete grade mask for each road segment. Specifically, the predicted linguistic grade from GPT-v is aligned with the categorical grade space (high / medium / low), while the VLM prior provides soft scores for each grade conditioned on the image–prompt pair. A lightweight prediction head takes as input the concatenated geometric features, the VLM grade scores, and an embedding of the GPT-v textual output, and outputs the final grade for each skeleton segment. The segment-wise grades are then projected back to the image space along the reconstructed skeletons and dilated to cover the full road width, producing a dense grade mask in which each pixel belonging to the road network is color-coded according to its hierarchical class. In this way, T-HRN transforms low-level geometric measurements into language-guided, semantically meaningful road grades, enabling interpretable and flexible hierarchical road network understanding.

\begin{table}[!t]
    \centering
    \caption{Summary of the quantitative results on SYSU-HiRoad dataset. The best results are in \textbf{bold}. (\%)}
    \begin{tabular}{c|cccc}
        \toprule
        & Recall & Precision & IoU & F1-score \\
        \midrule
        LinkNet & 69.66 & 72.66 & 54.56 & 68.89 \\
        D-LinkNet & 70.70 & 70.84 & 54.08 & 68.67 \\
        SGCNNet & 64.63 & \textbf{73.06} & 51.71 & 66.07 \\
        RCFSNet & 74.43 & 71.30 & 57.07 & 71.22 \\
        MACUNet & 72.16 & 72.60 & 56.64 & 70.94 \\
        FRCFNet &70.22&71.34 &54.17 &68.70 \\
        \midrule
        \rowcolor{gray!20}
        \textbf{Ours w/o ARR.} & 73.46 & 71.89 & 57.16 & 71.21 \\
        \rowcolor{gray!40}
        \textbf{Ours w/ ARR.} & \textbf{74.50} & 72.62 & \textbf{58.21} & \textbf{72.14} \\
        \bottomrule
    \end{tabular}
    \label{Table: Summary of the quantitative results}
\end{table}

\section{Experiments}

\subsection{Experiment Settings}

All experiments are implemented on an NVIDIA RTX 3090 Ti. The batch size is set to 2, and Adam is used as the optimizer with an initial learning rate of 2e-4. A step decay strategy is adopted for updating the learning rate: if the loss function does not decrease over 3 epochs, the learning rate is reduced by a factor of 5; training terminates if the learning rate falls below 5e-7. In addition, data augmentation is employed to enhance the model's generalization ability, which mainly includes random adjustments of hue, saturation, and brightness, random translations, scaling, rotations, and random horizontal and vertical flips. To evaluate performance, Accuracy, Precision, Recall, mIoU, and F1-score are used as evaluation metrics.

\subsection{Road Extraction Performance}

To comprehensively evaluate the performance of different methods in road segmentation tasks, we conduct extensive comparative experiments. We compared the proposed FORCE-Net on the SYSU-HiRoads dataset with several classic road segmentation methods, including LinkNet \citep{Chaurasia2017LinkNet}, D-LinkNet \citep{Zhou2018DLinkNetLW}, SGCNNet \citep{Zhou2022SplitD}, RCFSNet \citep{Yang2023GeoscienceR}, and MACUNet \citep{Li2022MACU}. For fairness and reproducibility, we include only methods with publicly available implementations, and all results are obtained by testing on our local machine.

\subsubsection{SYSU-HiRoads Dataset Evaluation} 

\begin{figure*}[!t]
    \centering
    \includegraphics[width=1\linewidth]{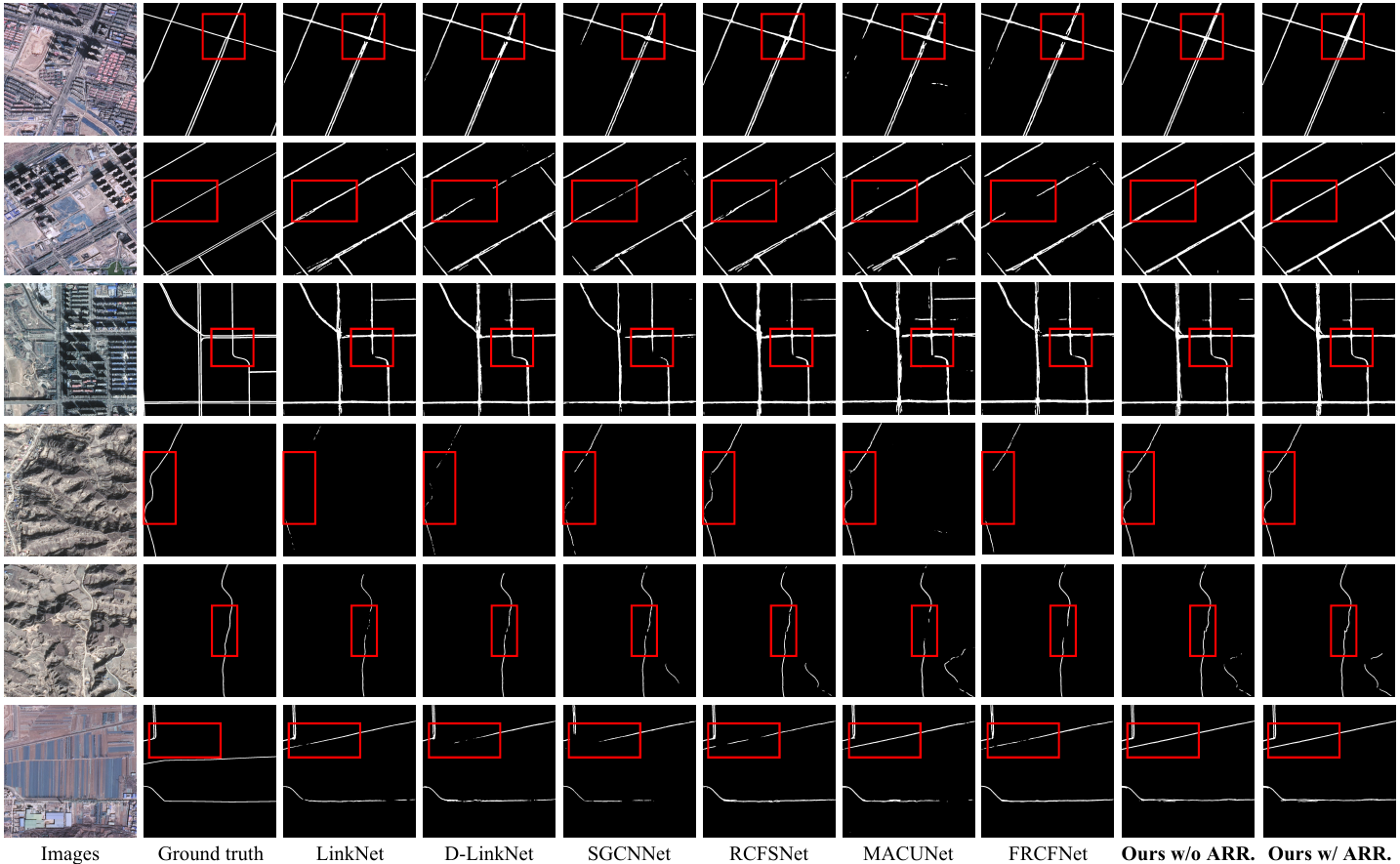}
    \caption{Visualizations of our FORCE-Net compared with different methods on SYSU-HiRoads dataset. }
    \label{fig: results}
\end{figure*}

Tab.~\ref{Table: Summary of the quantitative results} shows the quantitative results on the SYSU-HiRoad dataset. Our method achieves the best performance in terms of Recall, IoU, and F1-score. Even without using road reconstruction, the proposed network still exhibits competitive performance.

Fig.~\ref{fig: results} visually illustrates the segmentation results of different methods. In urban scenes, our method accurately identifies shadowed areas, while other methods fail to achieve this. In addition, the roads segmented by our method are more complete, with smoother edges and better continuity. Moreover, mountainous regions contain many rural roads, and the scattered distribution of buildings along these roads introduces significant noise, complicating road extraction. Nevertheless, even in such complex mountainous environments, our method still achieves excellent segmentation performance.

\subsubsection{CHN6-CUG Dataset Evaluation} 

\begin{table}[!t]
    \centering
    \caption{Summary of the quantitative results on CHN6-CUG dataset. The best results are in \textbf{bold}. (\%)}
    \begin{tabular}{c|cccc}
        \toprule
        & Recall & Precision & IoU & F1-score \\
        \midrule
        LinkNet & 37.39&73.03 &33.93 & 44.97\\
        D-LinkNet &58.75 & 72.76& 50.02 &61.94\\
        SGCNNet &52.42 & \textbf{74.70} &46.36 &58.06\\
        RCFSNet &\textbf{63.25} &62.47 &46.79 &58.67\\
        MACUNet &52.35 &69.67 &43.26 &55.29 \\
        FRCFNet & 58.72& 71.36 &49.28 &61.17\\
        \midrule
        \rowcolor{gray!20}
        \textbf{Ours w/o ARR.} &60.39 &74.30 &51.31 &63.20\\
        \rowcolor{gray!40}
        \textbf{Ours w/ ARR.} & 61.45 &73.72 &\textbf{51.81} & \textbf{63.89}\\
        \bottomrule
    \end{tabular}
    \label{Table: Summary of the quantitative results_CHN6}
\end{table}

\begin{figure*}[!t]
    \centering
    \includegraphics[width=1\linewidth]{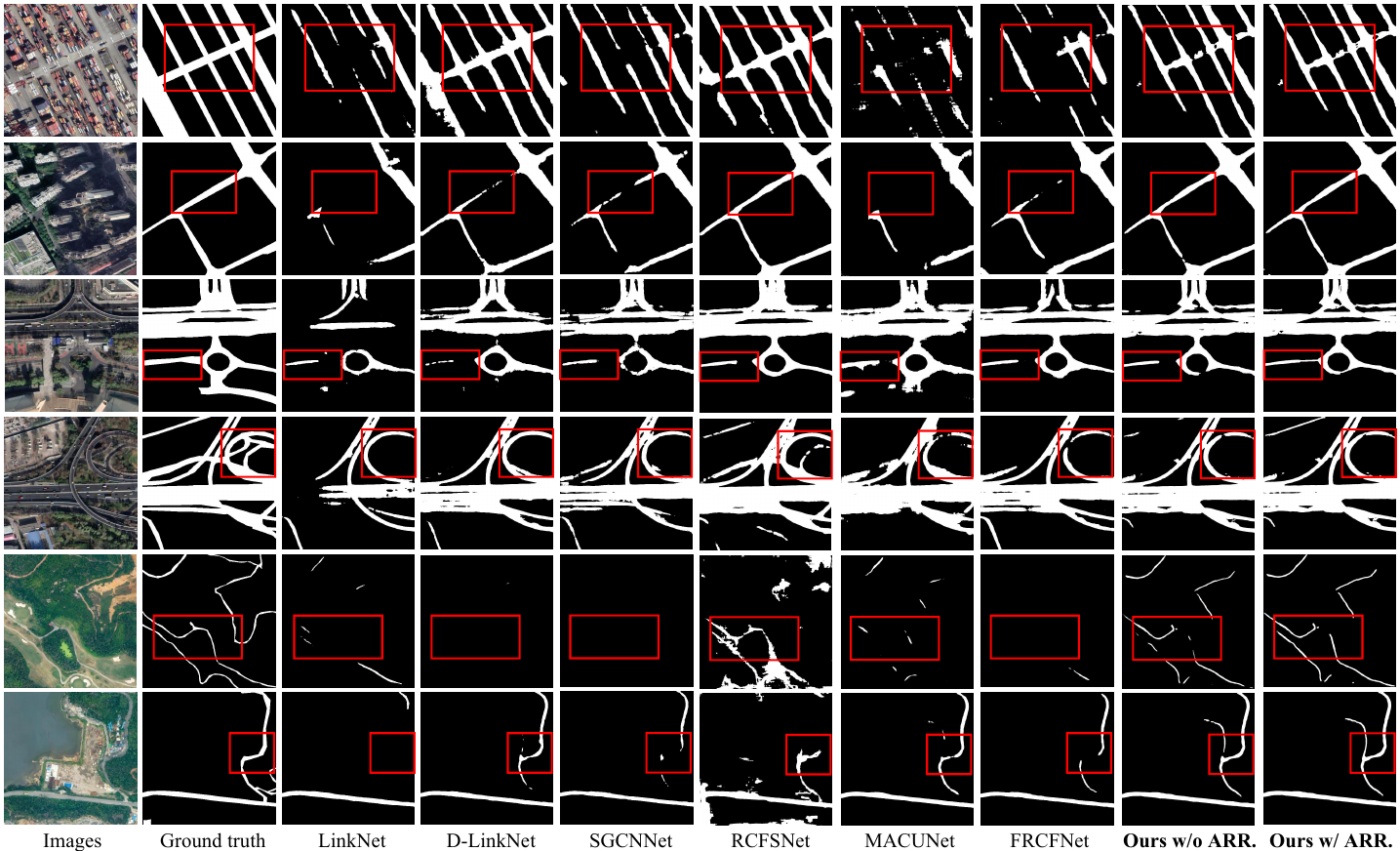}
    \caption{Visualizations of our FORCE-Net compared with different methods on CHN6-CUG Road dataset.}
    \label{fig: CHN6-CUG_results}
\end{figure*}

Tab.~\ref{Table: Summary of the quantitative results_CHN6} and Fig.~\ref{fig: CHN6-CUG_results} present quantitative comparison results and the visualized predictions of FORCE-Net against several classic road segmentation models on the CHN6-CUG dataset.

FORCE-Net demonstrates a stronger capability in recognizing linear features, particularly in challenging areas such as open docklands and regions with tree occlusions (See Rows \#1 and \#3 of Fig.~\ref{fig: CHN6-CUG_results}). Additionally, the results in Row \#2 indicate that FORCE-Net outperforms other models in identifying road surfaces under shadow. When it comes to curvy and narrow mountain roads (Rows \#5 and \#6), the proposed method exhibits superior recognition performance. However, as shown in Row \#4, none of the models successfully detected roads in more complex ramp scenarios with higher curvature and structural intricacy.

Moreover, FORCE-Net achieves the highest IoU score of 51.81\% and the highest F1 score of 63.89\% on the CHN6-CUG dataset, surpassing the second-best RCFSNet by 5.02\% and 5.22\%, respectively. These results clearly demonstrate the superiority of our design. Furthermore, since FORCE-Net already provides a relatively complete initial prediction of road structures on the CHN6-CUG dataset, applying an automatic road reconstruction method does not lead to significant improvements in accuracy.

\subsubsection{Ablation Study for FORCE-Net} 

\begin{table*}
    \centering
    \caption{Results for ablation study on SYSU-HiRoads Dataset, where the best results are in \textbf{bold}. (\%)}
    \begin{tabular}{ccccccc}
        \toprule
        Baseline & PMSE & FDE & Recall & Precision & IoU & F1-score \\
        \midrule
        \checkmark & & & 67.48 & 68.73& 50.79 &65.50\\
        \checkmark & \checkmark & & 71.57 & 70.13& 54.48 & 68.98\\
        \checkmark & & \checkmark & \textbf{75.16} &68.39 & 55.24 & 69.81 \\
        \checkmark & \checkmark & \checkmark & 73.46 & \textbf{71.89} & \textbf{57.16} & \textbf{71.21} \\
        \bottomrule
    \end{tabular}
    \label{Table:Results for ablation study on Multi-Grade Road Dataset}
\end{table*}

\begin{figure}[]
    \centering
    \includegraphics[width=0.9\linewidth]{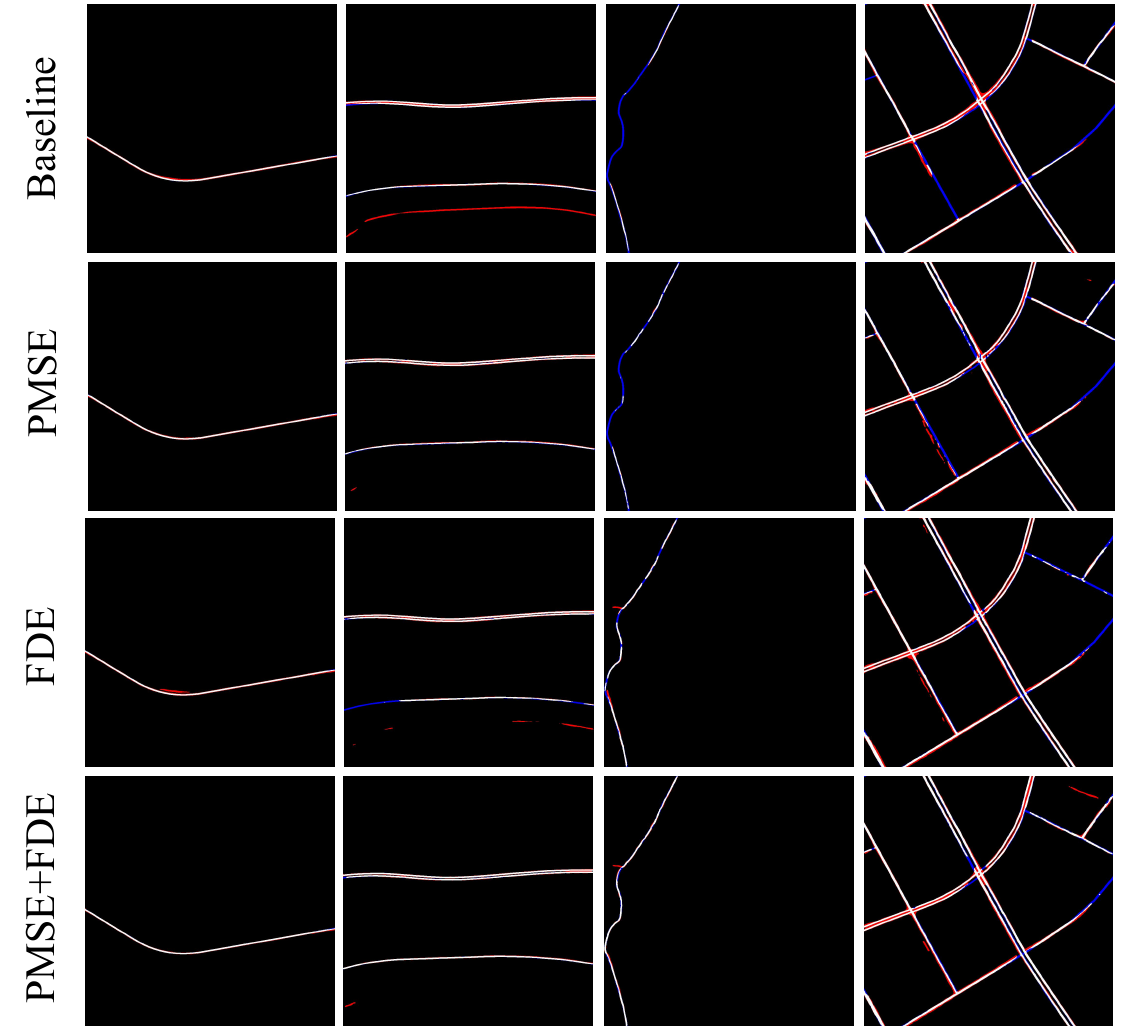}
    \caption{Visualization of ablation study for FORCE-Net. The White, Red, and Blue masks denote the true positive, false positive, and false negative pixels, respectively.}
    \label{fig: Visualization of ablation study}
\end{figure}

\begin{figure}[]
    \centering
    \includegraphics[width=0.9\linewidth]{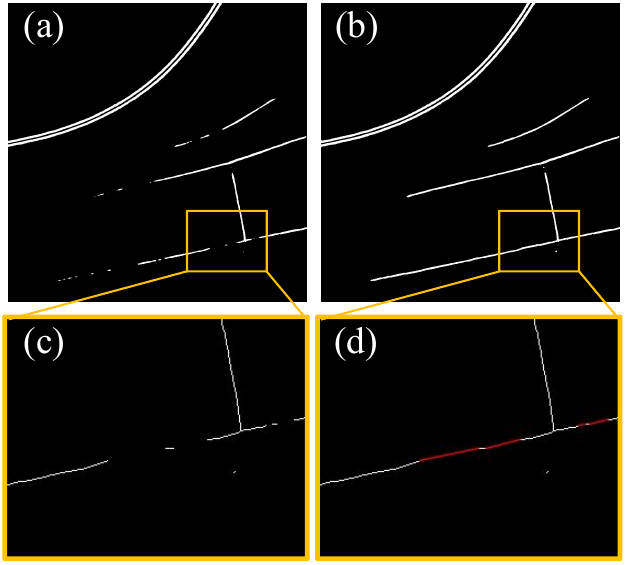}
    \caption{Display of the reconstruction process of the centerline.}
    \label{fig: reconstruction}
\end{figure}

To obtain more complete road information, we propose FDE and PMSE to capture the long-range relationships within road regions. To verify the effectiveness of these modules, we conduct experiments, as shown in Tab.~\ref{Table:Results for ablation study on Multi-Grade Road Dataset}. The baseline is a U-Net architecture based on ResNet-34. When PMSE is added to the baseline, the IoU improves from 50.79\% to 54.48\%, and the F1-score increases from 65.50\% to 68.98\%. When FDE is added to the baseline, the IoU score increases by 4.45\% and the F1-score by 4.31\%. When both PMSE and FDE are jointly integrated into the network, the IoU and F1-score improve by 6.37\% and 5.71\%, respectively.

From Fig.~\ref{fig: Visualization of ablation study}, it can be observed that the FDE module performs excellently in predicting long-distance roads, accurately outlining their contours. This fully demonstrates the positive role of the Fourier operator in road extraction tasks, effectively capturing long-distance road features. However, this module is relatively sensitive to the surrounding environment and can be easily interfered with by nearby objects during identification, leading to confusion between roads and surrounding features such as pipelines, buildings, or bare land, which reduces the accuracy of road extraction. 

On the other hand, PMSE shows outstanding advantages in reducing interference from surrounding objects. This indicates that the parallel structure of dilated convolutions helps to filter out distracting information and focus on the feature extraction of the road itself. Due to the complementary effects of these two modules, our method not only effectively captures road features but also improves the issue of fragmented road predictions, achieving accurate and complete road feature extraction.

The visual comparison before and after reconstruction is presented in Fig.~\ref{fig: reconstruction}. In the segmentation image before reconstruction, the fractured roads exhibit obvious discontinuities, with many missing and fragmented areas in the road regions. In contrast, after reconstruction, the integrity of the roads is significantly enhanced, and the originally broken segments are effectively connected and supplemented. This results in a more coherent and accurately segmented region that precisely reflects the extension direction and trajectory of the roads.

\subsubsection{Ablation Study for T-HRN} 

\begin{table}[!t]
\centering
\caption{Summary of geometric descriptors used in T-HRN. The $\checkmark$ indicates whether the descriptor is used in the final configuration.}
\begin{tabular}{c|c|c|c}
\toprule
Category & Descriptors & Symbol & Used \\
\midrule
Scale   & Segment length          & $L$         & \checkmark \\
Scale   & Mean road width         & $W$         & \checkmark \\
\midrule
Shape   & Straightness            & $S$         & \checkmark \\
Shape   & Mean curvature          & $C$         & \checkmark \\
Shape   & Orientation variability & $O$         & -- \\
\midrule
Context & Node degree (junctions) & $D$         & \checkmark \\
Context & Local road density      & $\rho$      & \checkmark \\
Context & Lane-count              & $L_{c}$     & --  \\
\bottomrule
\end{tabular}
\label{tab:geom_features}
\end{table}

\begin{table}[!t]
\centering
\caption{Ablation study of T-HRN on SYSU-HiRoads. G: geometry, I: image features, V: CLIP VLM prior, L: GPT-v language output. The best results are in \textbf{bold}. (\%)}
\begin{tabular}{c|cccccc}
\toprule
Variant & OA & $\kappa$ & mIoU & wF1 & mDice & SegAcc \\
\midrule
G & 66.8 & 54.1 & 41.6 & 58.2 & 50.9 & 52.3 \\
G + I & 68.3 & 55.9 & 43.1 & 59.6 & 52.3 & 54.0 \\
G + V & 71.9 & 60.5 & 48.1 & 63.1 & 56.9 & 58.9 \\
G + L & 67.5 & 55.0 & 42.2 & 58.9 & 51.5 & 53.2 \\
\textbf{Full T-HRN} & \textbf{72.6} & \textbf{61.3} & \textbf{48.9} & \textbf{64.2} & \textbf{57.8} & \textbf{60.5} \\
\bottomrule
\end{tabular}
\label{tab:ablation-thrn}
\end{table}

\textbf{Evaluation Metrics.} We evaluate the proposed method at both the pixel level and the road-segment level in order to assess not only local classification quality but also the consistency of hierarchical grading at the object level.

\begin{table*}[!t]
\centering
\caption{Ablation study of VLM in T-HRN in terms of road-grade prediction on SYSU-HiRoads. The best results are in \textbf{bold}. (\%)}
\begin{tabular}{c|ccc|ccc|ccc}
\toprule
\multirow{2}{*}{Variant} 
    & \multicolumn{3}{c|}{SegAcc} 
    & \multicolumn{3}{c|}{F1} 
    & \multicolumn{3}{c}{IoU} \\
\cmidrule(lr){2-4} \cmidrule(lr){5-7} \cmidrule(lr){8-10}
    & H & M & L 
    & H & M & L 
    & H & M & L \\
\midrule
CLIP-B/32 \citep{radford2021learning} & 16.7 & 39.3 & 85.7 & 14.3 & 17.9 & 59.0 & 7.7 & 9.9 & 41.8 \\
CLIP-L/14 \citep{radford2021learning} & 50.0 & 70.4 & 61.1 & 46.8 & 63.2 & 58.7 & 42.0 & 50.6 & 54.2 \\
DINOv2-ViT-B \citep{oquab2023dinov2} & 58.3 & 76.5 & 88.9 & 55.9 & 71.8 & 82.7 & 38.8 & 56.0 & 70.6 \\
BLIP-Base \citep{li2022blip} & 44.1 & 62.7 & 73.8 & 40.7 & 58.6 & 69.9 & 25.6 & 41.4 & 53.8 \\
SigLIP-Base \citep{zhai2023sigmoid} & 47.5 & 66.0 & 78.4 & 44.3 & 61.5 & 74.8 & 28.4 & 44.4 & 59.7 \\
SigLIP-Large \citep{zhai2023sigmoid} & 49.2 & 68.1 & 80.3 & 46.9 & 63.4 & 76.9 & 30.7 & 46.4 & 62.4 \\
OpenCLIP \citep{ilharco2021openclip} & 42.6 & 61.2 & 75.0 & 39.1 & 56.8 & 71.1 & 24.3 & 39.7 & 55.2 \\
X-CLIP \citep{ma2022x} & 45.8 & 64.0 & 77.6 & 42.0 & 59.7 & 73.6 & 26.6 & 42.5 & 58.2 \\
ALBEF \citep{li2021align} & 18.6 & 51.4 & 72.8 & 17.1 & 45.9 & 66.2 & 9.4 & 30.0 & 49.5 \\
EVA02 \citep{fang2024eva} & 51.0 & 69.7 & 82.1 & 48.1 & 65.1 & 78.5 & 31.8 & 48.2 & 64.6 \\
\bottomrule
\end{tabular}
\label{tab:thrn_grade_metric_ablation}
\end{table*}

Pixel-level metrics are several standard metrics computed over all road pixels. For each class, we compute precision, recall, F1-score, and Intersection-over-Union (IoU). The overall accuracy (OA) is defined as the proportion of correctly classified road pixels. The mean IoU (mIoU) is obtained by averaging class-wise IoUs across the three road grades. To quantify the overall agreement beyond chance, we further compute Cohen’s Kappa coefficient. Finally, we report the Sørensen–Dice coefficient for each class and its average across classes, which provides an additional overlap-based measure complementary to IoU.
 
To evaluate the correctness of hierarchical grading at the object level, we perform a road-segment-level assessment based on connected component analysis. For every predicted segment, we locate its corresponding ground-truth segment via the segment centroid and form pairs of predicted and true grades at the segment level. On this basis, we compute the overall segment-level grading accuracy, i.e., the proportion of road segments whose predicted grade matches the ground-truth grade. 

\textbf{Geometric Descriptors.} As illustrated in Table~\ref{tab:geom_features}, we design a compact set of geometric descriptors to characterize each road segment in terms of scale, shape, and network context. For scale, we use the segment length $L$ and the mean road width $W$, which jointly capture how long and how wide a segment is. These two cues are strongly correlated with the functional class of a road, as high-grade roads tend to be wider and span longer distances than low-grade roads.

In terms of shape, we retain straightness $S$ and mean curvature $C$ to describe whether a segment is predominantly straight or meandering. High-grade roads are usually designed to be straighter with gentle curvature, whereas low-grade roads in residential and rural areas often exhibit larger changes in curvature and irregular shapes. We also experiment with an orientation-variability descriptor $O$, which measures the variation of local tangent directions along the segment, but we find it to be largely redundant once $S$ and $C$ are included.

For network context, we incorporate node degree $D$ at junctions and local road density $\rho$ around each segment. Node degree reflects how a segment is embedded in the road graph, while local density captures whether the segment lies within a dense local street fabric and a sparse arterial network. We further consider a lane-count proxy $L_c$ derived from the estimated road width and typical lane widths. However, in practice, it does not produce consistent gains beyond $W$ and is thus excluded from the final T-HRN configuration.

\textbf{Model Configurations.} Tab.~\ref{tab:ablation-thrn} reports the ablation results of T-HRN on SYSU-HiRoads by progressively injecting different features into the geometry-based baseline. Incorporating the CLIP-based VLM prior yields the most substantial gain across all metrics, boosting OA from 66.8\% to 71.9\%, mIoU from 41.6\% to 48.1\%, and SegAcc from 52.3\% to 58.9\%. This suggests that the VLM prior provides strong semantic guidance for road-grade discrimination beyond geometry and appearance alone. The complete model achieves the best overall performance, demonstrating that language output further refines the VLM-enhanced representation, while the VLM prior remains the primary contributor, and geometric features serve as the essential foundation.

\textbf{VLM Configurations.} Tab.~\ref{tab:thrn_grade_metric_ablation} investigates the effect of different VLM backbones in T-HRN for road-grade prediction. Stronger VLMs lead to more robust grade recognition. The vanilla models show limited capability, suggesting imbalanced behavior towards easier categories. Among all candidates, DINOv2-ViT-B achieves the best performance, particularly on low-grade roads, and also improves medium-grade recognition, indicating a stronger representation capacity for fine-grained linear structures. EVA02 is competitive, whereas others provide moderate gains but remain inferior to DINOv2, especially on dominant roads where confusion with adjacent classes is more frequent. These results verify that the choice of VLM backbone is a critical factor in T-HRN, and higher-quality visual priors translate into better grade-level segmentation and segment-wise consistency.

\begin{figure*}
    \centering
    \includegraphics[width=\linewidth]{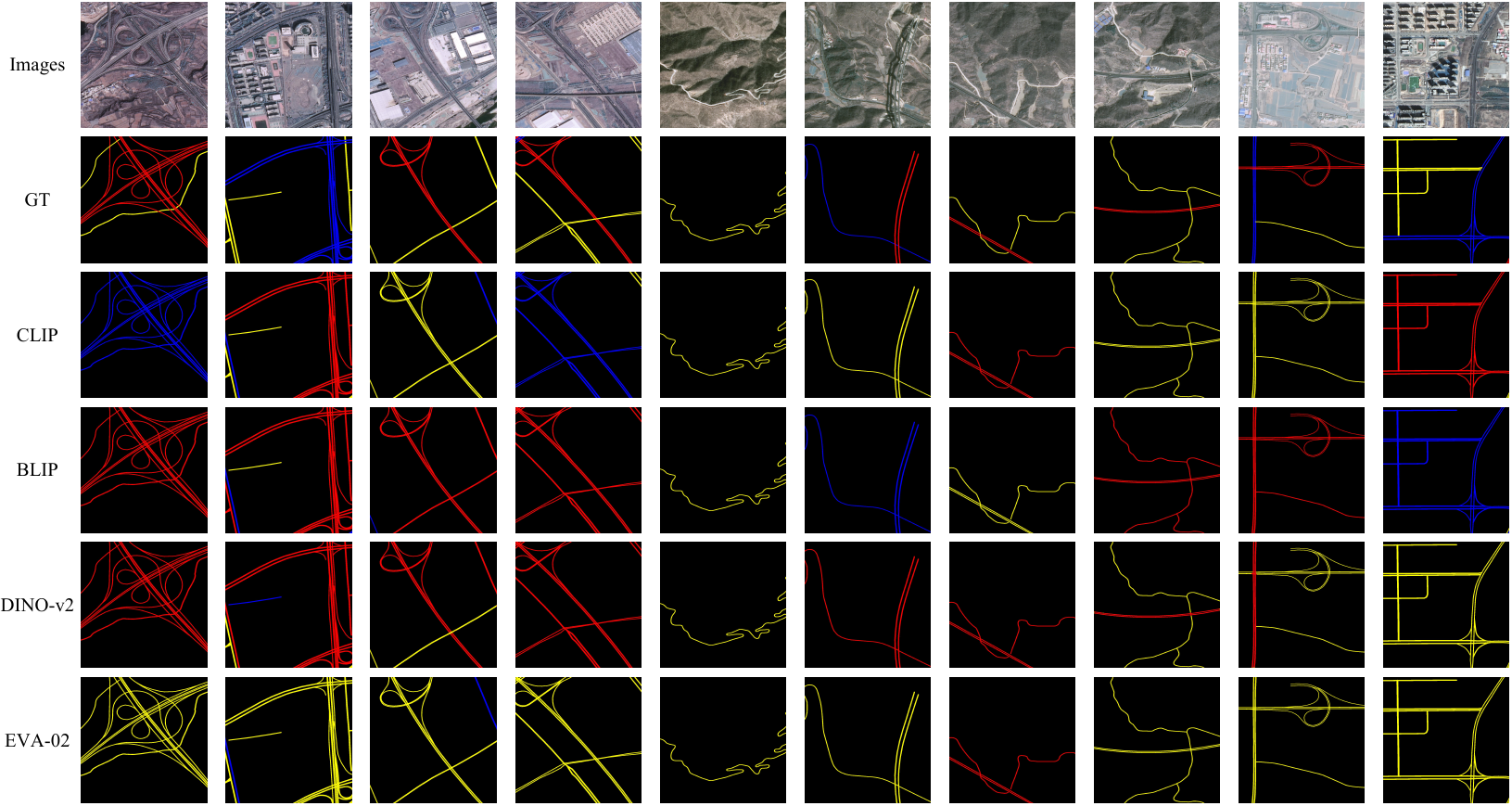}
    \caption{Visualizations of our RoadReasoner on representative cases of SYSU-HiRoads dataset. The red, blue, and yellow represent the high, medium, and low grade roads, respectively.}
    \label{fig: multi-grade-pred-results}
\end{figure*}

As shown in Fig.~\ref{fig: multi-grade-pred-results}, we qualitatively compare different VLM in RoadReasoner on representative SYSU-HiRoads scenes. CLIP and BLIP-based variants are more prone to grade confusion in complex intersections and dense networks, leading to missed segments. DINO-v2 yields the most accurate results, showing that slight grade ambiguities remain in a few challenging layouts. These visual comparisons corroborate the quantitative results in Tab.~\ref{tab:thrn_grade_metric_ablation}, indicating that stronger VLM priors improve grade-level consistency for road-grade prediction.

\section{Challenges and Applications}\label{Sec: Challenges}

\subsection{Challenges}

Although SYSU-HiRoads and the proposed RoadReasoner framework significantly advance automatic road extraction and fine-grained hierarchy classification from high-resolution remote sensing imagery, several open challenges remain and point to promising directions for future research.

\textbf{(1) Generalization across regions, sensors, and temporal domains.}
SYSU-HiRoads is constructed from GF-2 imagery over Henan Province, China, and therefore reflects the imaging characteristics, urban morphology, and road design standards of this particular context. In practice, road appearance can vary drastically across countries and climate zones due to differences in construction materials, lane markings, land-cover patterns, and urban form. Moreover, imagery from different sensors (e.g., WorldView, Sentinel-2, UAV) exhibits diverse resolutions, spectral characteristics, and noise patterns, while temporal changes, such as newly built or demolished roads, introduce additional distribution shifts. How to make road extraction and hierarchy classification models robust to such spatial, temporal, and sensor-domain variations remains a critical challenge for deploying RoadReasoner-like systems on a global scale.

\textbf{(2) Ambiguity and inconsistency in hierarchical road labels.}
Although SYSU-HiRoads provides three-level hierarchical labels by integrating national administrative standards, OSM attributes, and expert interpretation, road hierarchy is inherently context-dependent. For example, a road classified as a “high-grade” artery in a rural area may function more like a “medium-grade” road in a dense metropolitan core. Administrative classifications, functional roles, and observed traffic volumes do not always align perfectly. This can lead to label ambiguity, edge cases between classes, and even inter-annotator disagreement. Building models that can gracefully handle such ambiguity, explicitly represent label uncertainty, and remain robust to moderate inconsistencies in training data is still an open research problem.

\textbf{(3) Reliability and controllability of language–vision models.}
T-HRN leverages a VLM and GPT-v to infer road grades from geometry-aware prompts and visual context. While this yields interpretable predictions, it also introduces new challenges related to reliability, stability, and controllability. Vision–language models can be sensitive to prompt design, exhibit hallucinations, or over-rely on spurious correlations, especially when confronted with remote sensing imagery that differs from their pre-training distribution. Ensuring that language-based decisions faithfully reflect geometric and contextual evidence, preventing unintended biases in grade assignment, and developing mechanisms to calibrate or verify VLM outputs (e.g., via consistency checks, uncertainty estimation, or human-in-the-loop verification) are important directions for future work.

\subsection{Applications}

Despite these challenges, SYSU-HiRoads and RoadReasoner open up multiple applications in remote sensing, transportation planning, and urban studies.

\textbf{(1) Automated mapping and high-definition base maps.}
High-quality road maps are a cornerstone of many cartographic and GIS products. By jointly providing pixel-level road masks, vector centerlines, and hierarchical labels, SYSU-HiRoads can serve as a strong supervision source for training models that support automated or semi-automated map updating. RoadReasoner can be used to refine existing road layers by improving the completeness of road extraction, repairing broken centerlines, and assigning functionally meaningful grades. This is valuable for producing high-definition base maps that underpin navigation, logistics, and urban digital twins.

\textbf{(2) Transport planning, accessibility, and equity analysis.}
Road hierarchy is closely linked to capacity, speed limits, and connectivity; therefore, it plays a central role in transport planning and accessibility assessment. Accurate fine-grained road grades allow planners to better estimate travel times, identify bottlenecks, and evaluate the distribution of high-capacity corridors across different neighborhoods. Combined with population, land-use, and socioeconomic data, outputs from RoadReasoner can support studies of spatial accessibility to jobs, healthcare, and education and inform policies aimed at improving transport equity and mitigating infrastructure-induced inequalities.

\textbf{(3) Disaster response and resilience assessment.}
In disasters such as floods, earthquakes, or landslides, the functional hierarchy of roads strongly influences evacuation routes, emergency service access, and logistics operations. By automatically extracting road networks and their grades before and after an event, RoadReasoner can help identify critical links whose failure would severely disrupt connectivity, as well as redundancy in the network. When combined with change detection or damage assessment models, SYSU-HiRoads-style labels can be used to prioritize inspections and repairs on high-grade roads, thereby supporting rapid response and long-term resilience planning.

\textbf{(4) Support for autonomous driving and intelligent transportation systems.}
Autonomous vehicles and intelligent transportation systems require detailed knowledge of road geometry and hierarchy to plan safe and efficient trajectories. While many current high-definition maps rely on LiDAR and in situ surveys, high-resolution satellite imagery offers a complementary and potentially more scalable data source, especially in areas with limited ground infrastructure. RoadReasoner’s ability to recover thin, continuous road structures and assign grades can assist in bootstrapping or updating HD maps, particularly at the pre-survey stage or in regions where existing road databases are outdated or incomplete.

\section{Conclusion}\label{Sec: Conclusion}

In this study, we addressed two intertwined limitations in current remote sensing road infrastructure management research: the lack of datasets that jointly support road surface extraction, road network reconstruction, and hierarchy classification, and the under-explored potential of vision–language models for structured road grade understanding. To this end, we introduced \textbf{SYSU-HiRoads}, a large-scale hierarchical road dataset over Henan Province, China, and \textbf{RoadReasoner}, a vision–language–geometry framework that exploits SYSU-HiRoads to perform automatic road extraction and fine-grained hierarchy classification from high-resolution satellite imagery.

SYSU-HiRoads provides pixel-level road masks, vector-level centerlines, and three-level hierarchy labels, bridging the gap between segmentation-oriented and network-oriented road benchmarks. Moreover, we developed \textbf{FORCE-Net}, which improves the completeness and connectivity of road predictions by combining Fourier-based frequency domain enhancement, parallel multi-scale feature extraction, and automatic centerline reconstruction. We further proposed \textbf{T-HRN}, a text-guided hierarchical system that translates geometric descriptors of road segments into language prompts and queries a VLM and GPT-v to infer interpretable grade assignments. These components form \textbf{RoadReasoner}, which outputs dense pixel-wise road masks enriched with semantically meaningful hierarchical information.

Extensive experiments on SYSU-HiRoads and CHN6-CUG demonstrate that FORCE-Net is competitive with state-of-the-art road extraction methods, particularly in challenging scenarios involving thin, fragmented, and occluded roads. At the same time, RoadReasoner yields accurate and coherent road hierarchy maps, showing that coupling geometry-aware prompts with vision–language models is a viable pathway toward more interpretable and functionally informed road mapping. Beyond numerical gains, the proposed dataset and framework highlight how remote sensing, geometric reasoning, and foundation models can be integrated to move from “where are the roads” to “what roles do these roads play” within the network.

Looking ahead, this work provides a foundation for future research directions. SYSU-HiRoads and RoadReasoner can support the study of domain generalization across regions, the extension of language-guided hierarchy reasoning to other infrastructure networks such as railways, rivers, and utility corridors, and the development of human-in-the-loop systems in which domain experts and multimodal models jointly update and refine infrastructure information. More broadly, the proposed dataset and framework offer a practical basis for hierarchy-aware transport infrastructure mapping from Earth observation data, with potential value for road inventory updating, digital infrastructure management, and related built environment applications.

\section{Declaration of Competing Interest}

The authors declare that they have no known competing financial interests or personal relationships that could have appeared to influence the work reported in this paper.

\section{Data statement}

The SYSU-HiRoads dataset reconstructed in this study is available at https://doi.org/10.5281/zenodo.18642747 \citep{han_2026_18642747}.

\section{Acknowledgments}

We appreciate the valuable comments and constructive suggestions from the anonymous reviewers that helped improve the manuscript.

\bibliographystyle{cas-model2-names}
\bibliography{cas-refs}

\end{document}